\documentclass{article}
\usepackage[T1]{fontenc}
\usepackage{rtl_interact_preprint,times}

\usepackage{amsmath,amsfonts,bm}

\def\1{\bm{1}}

\DeclareMathAlphabet{\mathsfit}{\encodingdefault}{\sfdefault}{m}{sl}
\SetMathAlphabet{\mathsfit}{bold}{\encodingdefault}{\sfdefault}{bx}{n}

\usepackage{amsmath,amssymb}
\usepackage{booktabs,multirow,makecell,array}
\usepackage{graphicx,colortbl,float,microtype,pifont}
\usepackage{needspace}
\usepackage{wrapfig}
\usepackage{placeins}
\usepackage{hyperref,url,xspace}
\usepackage{tikz}
\usepackage{capt-of}
\usetikzlibrary{arrows.meta,positioning,calc,fit}
\usepackage{pgfplots}
\pgfplotsset{compat=1.18}
\definecolor{PaperBlue}{HTML}{3568B8}
\definecolor{PaperTeal}{HTML}{148F86}
\definecolor{PaperOrange}{HTML}{D97A24}
\definecolor{PaperRed}{HTML}{B94848}
\newcommand{\cmark}{\textcolor{PaperTeal}{\ding{51}}}
\newcommand{\xmark}{\textcolor{black!55}{\ding{55}}}

\newcommand{\method}{\textsc{RTL-Interact}\xspace}
\newcommand{\papertablestyle}{%
  \small
  \setlength{\tabcolsep}{4pt}%
  \renewcommand{\arraystretch}{1.12}}
\newsavebox{\papertablebox}
\newenvironment{papertablefit}{%
  \begin{lrbox}{\papertablebox}}{%
  \end{lrbox}%
  \ifdim\wd\papertablebox>\linewidth
    \resizebox{\linewidth}{!}{\usebox{\papertablebox}}%
  \else
    \usebox{\papertablebox}%
  \fi}
\newcommand{\pendinglabel}{Pending result---conditional draft text.}

\usepackage{listings}
\usepackage[most]{tcolorbox}
\tcbset{paperexample/.style={enhanced,breakable,colback=black!1,
  colframe=PaperBlue!65!black,colbacktitle=PaperBlue!9,coltitle=black,
  fonttitle=\small\bfseries,fontupper=\small,boxrule=0.5pt,arc=1.5mm,
  left=7pt,right=7pt,top=6pt,bottom=6pt,before skip=9pt,after skip=9pt,
  title after break={\textit{Continued}},pad at break*=2mm}}
\lstdefinestyle{paperliteral}{basicstyle=\ttfamily\footnotesize,
  columns=fullflexible,keepspaces=true,breaklines=true,breakatwhitespace=true,
  showstringspaces=false,aboveskip=4pt,belowskip=4pt,tabsize=2}

\title{RTL-Interact: Training EDA-Interactive\\Models for Multi-Task RTL Design}
\hypersetup{hidelinks,pdftitle={RTL-Interact: Training EDA-Interactive Models for Multi-Task RTL Design},pdfauthor={Wenxuan Shi; Yaoyu Zhu; Di Huang; Chongxiao Li; Xiaoyun Zhang; Muxin Song; Yutong Wu; Ziyuan Nan; Pengwei Jin; Tian Zhi; Xishan Zhang; Rui Zhang; Xing Hu; Zidong Du; Qi Guo; Yunji Chen}}
\author{\parbox[t]{\dimexpr\textwidth-2\tabcolsep\relax}{\raggedright\normalfont
{\bfseries Wenxuan Shi\textsuperscript{1}, Yaoyu Zhu\textsuperscript{1}, Di Huang\textsuperscript{1}, Chongxiao Li\textsuperscript{1}, Xiaoyun Zhang\textsuperscript{2}\\
Muxin Song\textsuperscript{3}, Yutong Wu\textsuperscript{1}, Ziyuan Nan\textsuperscript{2}, Pengwei Jin\textsuperscript{1}, Tian Zhi\textsuperscript{1}, Xishan Zhang\textsuperscript{1}\\
Rui Zhang\textsuperscript{1}, Xing Hu\textsuperscript{1}, Zidong Du\textsuperscript{1}, Qi Guo\textsuperscript{1}, Yunji Chen\textsuperscript{1}}\\
{\textsuperscript{1}Institute of Computing Technology, Chinese Academy of Sciences\\
\textsuperscript{2}University of Chinese Academy of Sciences\\
\textsuperscript{3}ShanghaiTech University}
}}

\begin{document}

\raggedbottom
\maketitle
\begin{abstract}

Large language models offer a promising route to register-transfer-level (RTL) design, but reliable implementations require checking and revision based on electronic design automation (EDA) tools' feedback. Existing RTL-specialized models often lack learned interaction with such EDA tools or rely on separate generation and repair components. We introduce \method, a multi-task post-training framework that trains a single RTL-specialized model to generate RTL, construct task-appropriate EDA checks, and revise implementations using tool feedback. The framework combines automatically constructed multi-task demonstrations with staged supervision from chain-of-thought to tool-integrated reasoning. We further introduce Pattern-Aware Policy Optimization (PAPO) for multi-task reinforcement learning. PAPO augments Group Relative Policy Optimization with correctness-gated rewards that favor concise successful checking while rewarding effective revision, and uses task grouping to preserve coverage across objectives. On CVDP, RTL-Interact-32B achieves 48.7\% weighted Pass@1 and 68.9\% Pass@5, surpassing its teacher model, DeepSeek-V3.2, while RTL-Interact-8B achieves 44.9\% weighted Pass@1. Additionally, we show that PAPO can reduce average tool calls by 63.8\% while improving Pass@1. Both models also outperform all RTL-specialized baselines on VerilogEval v2 and RTLLM v2, demonstrating the effectiveness of our method.

\end{abstract}
\section{Introduction}
\label{sec:intro}
\suppressfloats[t]
\begin{figure}[t]
\centering
\includegraphics[width=0.90\linewidth]{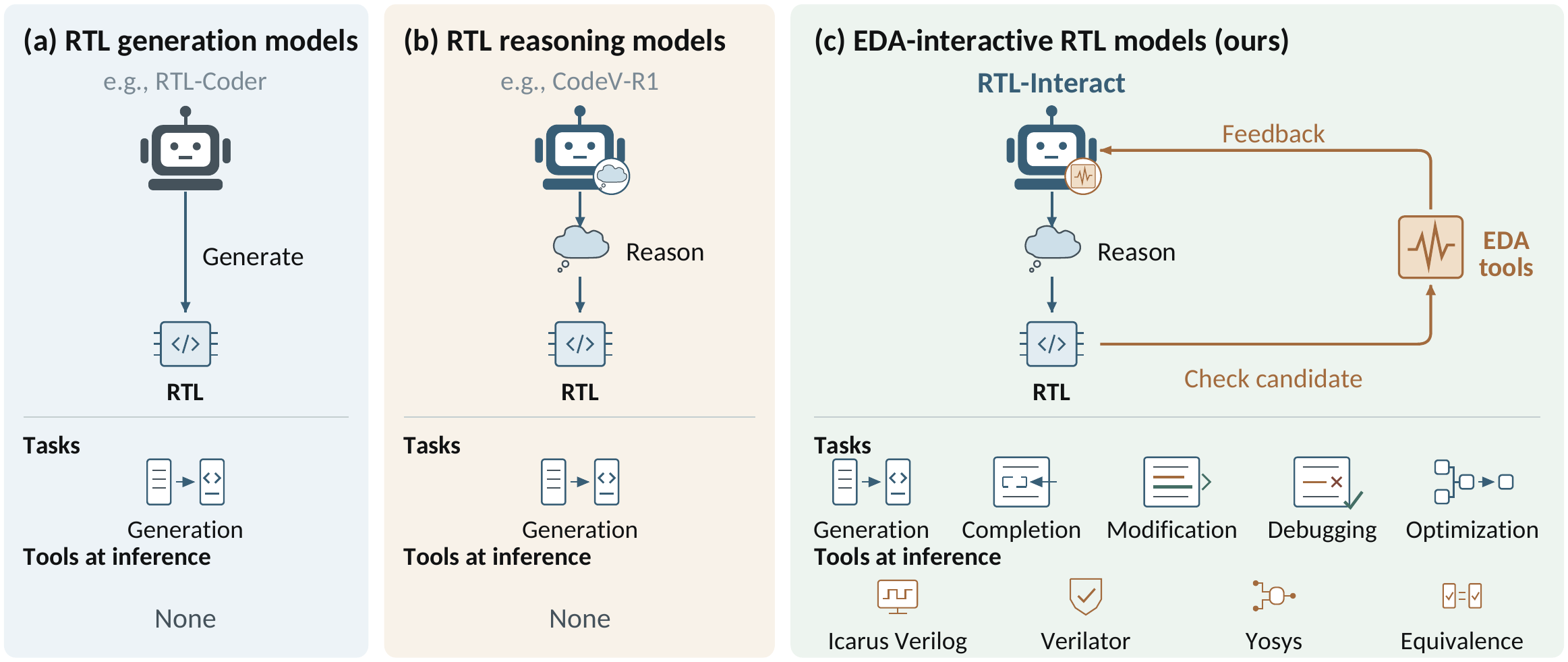}
\caption{Representative solution processes of (a) RTL generation models, such as RTL-Coder \citep{li2023rtlcoder}; (b) RTL reasoning models, such as CodeV-R1 \citep{zhu2025codevr1}; and (c) our EDA-interactive RTL model.}
\label{fig:rtl-model-comparison}
\end{figure}

Large language models (LLMs) offer a promising route to automating register-transfer-level (RTL) design.
Generating usable RTL requires not only syntactically valid code but also compliance with the functional requirements and implementation constraints stated in a specification.
Electronic design automation (EDA) tools provide feedback on whether candidate RTL compiles, behaves as intended in simulation, and meets task-specific design objectives.
Such feedback allows a model to check and revise its design before submitting a solution.

Existing RTL-specialized models improve code generation through domain supervision or EDA-derived training rewards, but do not explicitly learn to use EDA feedback to revise implementations during inference \citep{li2023rtlcoder,zhao2024codev,zhu2025codevr1}.
Tool-assisted workflows organize generation, checking, and revision through prompting and agent coordination, but rely on the underlying models' existing RTL generation and repair capabilities \citep{thakur2023autochip,zhao2024mage}.
Other approaches train models to revise RTL using EDA feedback, but require separate generation and repair adapters or cooperating models, adding deployment and coordination complexity \citep{cui2024origen,yu2026chipmate}.
These limitations motivate training a single RTL-specialized model for both RTL generation and feedback-based revision, without separate adapters or cooperating models.
Figure~\ref{fig:rtl-model-comparison} contrasts the solution processes of RTL generation and RTL reasoning models with those of EDA-interactive RTL models.
Achieving this across multiple RTL tasks requires learning task-appropriate checks and effective revisions while avoiding redundant tool calls.

To this end, we propose \method, a post-training framework for training a single RTL-specialized model to generate RTL, construct task-appropriate EDA checks, and revise candidate implementations using EDA feedback.
The framework consists of multi-task data construction, CoT-to-TIR staged supervision, and reinforcement learning (RL) with pattern-aware policy optimization.
Starting from the specification-RTL pairs in the open-source dataset, we automatically construct tasks with different code contexts and design objectives, then generate and filter teacher demonstrations using task-specific checks.
Each tool-interaction demonstration records the teacher's candidate implementations, EDA requests, returned feedback, subsequent revisions, and final implementation.

To develop both RTL reasoning and effective use of EDA feedback, we apply CoT-to-TIR staged supervision to improve RTL generation and revision using EDA feedback within the same model.
Tool-free chain-of-thought (CoT) examples and tool-integrated reasoning (TIR) demonstrations supervise different solution processes, making their organization an important training choice.
We first train on CoT examples from the open-source dataset, covering RTL syntax, hardware semantics, and reasoning from specifications to implementations.
We then train on multi-task TIR demonstrations to teach the model to construct EDA requests and use the returned feedback.
At a matched template-token budget, this staged approach improves CVDP \citep{pinckney2025cvdp} Pass@1 by 1.0\% over TIR-only supervision.
Separate evaluations also show improvements in both RTL generation and revision using EDA feedback over CoT-only supervision.

Multi-task RL further optimizes these behaviors through interaction with the EDA environment.
A reward based only on final correctness does not distinguish how a correct implementation was checked and revised.
We introduce Pattern-Aware Policy Optimization (PAPO), which incorporates correctness-gated rewards for EDA-checking patterns and task grouping into Group Relative Policy Optimization (GRPO) \citep{shao2024deepseekmath}.
For solutions that pass final verification, the reward favors concise, successful checking while retaining positive rewards for successful revision.
To maintain training coverage across different objectives, task grouping keeps prescribed task-group proportions among the samples retained for each policy update.
Across the full RL stage, the 8B model improves CVDP Pass@1 by 5.4\% while reducing the average number of tool calls by 63.8\%.

On CVDP, RTL-Interact-32B surpasses its tool-enabled teacher, DeepSeek-V3.2, achieving 48.7\% weighted Pass@1.
RTL-Interact-8B achieves 44.9\%, with both models outperforming other RTL-specialized models of comparable size.
These gains extend to VerilogEval v2 \citep{liu2023verilogeval,pinckney2024revisiting} and RTLLM v2, where both model sizes outperform other RTL-specialized baselines in Pass@1 across code completion and specification-to-RTL tasks.

Our contributions are as follows:
\begin{itemize}
\item We develop a multi-task post-training framework that uses task-specific teacher demonstrations and CoT-to-TIR staged supervision to train a single RTL-specialized model for RTL generation and revision using EDA feedback.
\item We introduce PAPO, which incorporates correctness-gated rewards for EDA-checking patterns and task grouping into GRPO to improve multi-task RTL performance with fewer tool calls and shorter response length, achieving joint gains in both RTL correctness and EDA interaction efficiency.
\item Our RTL-Interact-32B achieves 48.7\% Pass@1 and 68.9\% Pass@5 on CVDP, surpassing its teacher model DeepSeek-V3.2 on both metrics.
Our models also outperform all RTL-specialized baselines with large gains in Pass@1 on VerilogEval v2 and RTLLM v2.
We will release both models, together with the training data and code.
\end{itemize}

\section{Related Work}
\paragraph{RTL generation model training.}
RTLCoder and CodeV use domain data and supervised fine-tuning; BetterV adds discriminative guidance, and CraftRTL
constructs synthetic data and repair examples
\citep{li2023rtlcoder,zhao2024codev,pei2024betterv,liu2024craftrtl}.
CodeV-R1 combines reasoning distillation with testbench-based verifiable
rewards, while ChipSeek uses hierarchical EDA-derived rewards to optimize
functional correctness and hardware efficiency
\citep{zhu2025codevr1,chen2026chipseek}.
These methods primarily use EDA tools to evaluate generated designs.
Our \method instead learns to use EDA tools during inference to check and
revise candidates based on feedback before returning the final RTL, thereby
improving RTL task performance.

\paragraph{Tool-assisted RTL agents and workflows.}
AutoChip organizes iterative generation using compilation and simulation
feedback, while RTLFixer combines interactive repair with retrieved guidance
\citep{thakur2023autochip,tsai2024rtlfixer}.
MAGE, VerilogCoder, and LocalV organize generation and debugging through role
coordination, graph-based planning, and information locality, respectively
\citep{zhao2024mage,ho2025verilogcoder,lyu2026localv}.
ACE-RTL combines a specialized generator with a general-purpose reflector
\citep{deng2026acertl}.
These approaches primarily organize model capabilities through inference-time prompting and workflow design.
Our \method instead strengthens the model's own ability to generate RTL and revise it using EDA feedback through domain-specific training, complementing these inference-time approaches.

\paragraph{Training for tool interaction.}
ReAct interleaves reasoning and actions through prompting and
trajectory fine-tuning, Toolformer learns API use through
self-supervision, and ToRA trains reasoning models on tool-interaction
trajectories \citep{yao2023react,schick2023toolformer,gou2023tora}.
Within RTL design, OriGen trains separate generation and repair adapters to
correct compilation errors, while ChipMATE jointly trains Verilog and Python
reference-model agents for cross-verification
\citep{cui2024origen,yu2026chipmate}.
These methods learn feedback-based correction around compiler diagnostics or
cross-language behavioral comparison. 
Our \method instead trains a single model to construct task-appropriate EDA checks and revise RTL using their feedback across functional and optimization tasks, reducing the deployment overhead for multiple models or adapters.

\section{Method}
\label{sec:method}
\newcommand{\methodtablespresent}{1}

\subsection{Task Formulation and EDA Environment}
\label{sec:setup}

A task $x$ specifies hardware behavior and may include partial RTL or an existing design. A policy $\pi_\theta$ produces final RTL $y$ through a trajectory $\tau=(z_0,a_0,o_0,\ldots,z_L,y)$ with $L$ tool requests, where $z_\ell$, $a_\ell$, and $o_\ell$ denote model-generated reasoning, tool requests, and environment observations. Each request specifies a tool and arguments. The policy conditions on the task and history to reason, request checks, or return RTL satisfying the requirements and implementation constraints. With no requests, this reduces to direct RTL generation. Here $\ell$ indexes interactions, not tokens.

The environment exposes simulation, lint-checking, synthesis-analysis, and simulation-based equivalence-checking interfaces (Appendix~\ref{app:repro}, Table~\ref{tab:tool-interfaces}). Icarus Verilog, Verilator, and Yosys provide compilation/simulation, lint-checking, and synthesis functions \citep{icarusdocs,verilatordocs,yosysdocs}. Simulation-based equivalence checking (hereafter equivalence checking) uses CodeV-R1's rule-based testbench generator to compare candidate and reference outputs under shared stimuli \citep{zhu2025codevr1}. These testbenches serve only equivalence checking and are distinct from model-authored simulation testbenches. Testbenches and execution logs used for independent final verification are not supplied as model context.

\ifdefined\compactdraft\begin{figure}[H]\else\begin{figure}[t]\fi
\centering
\includegraphics[width=\linewidth]{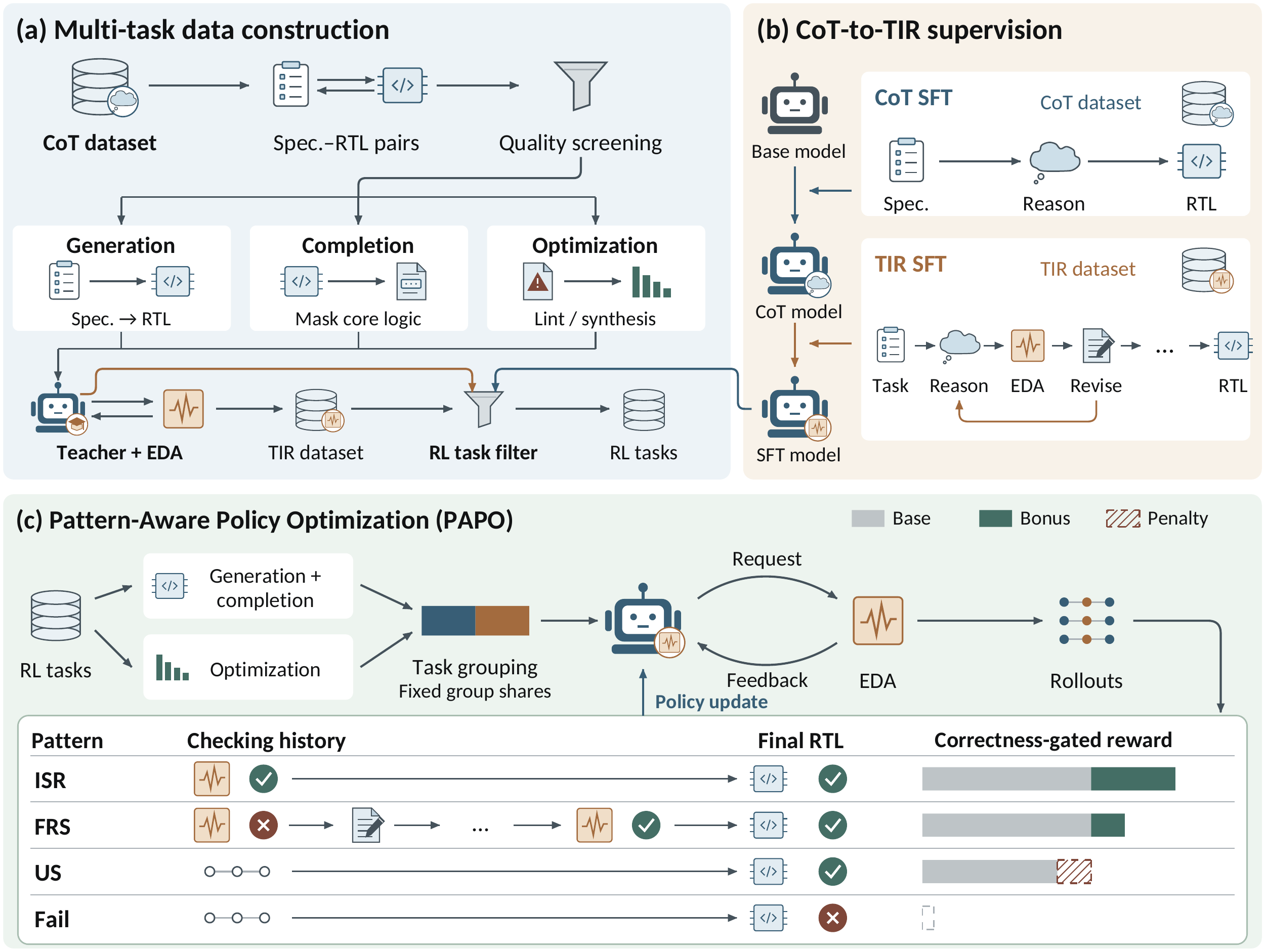}
\caption{Overview of \method:
(a) multi-task TIR data construction and RL task selection starting from the CoT dataset;
(b) CoT-to-TIR staged supervision; and
(c) PAPO with pattern-aware trajectory rewards and task grouping.
Reward bars are schematic; incorrect final RTL receives zero reward.}
\label{fig:overview}
\end{figure}

\subsection{Multi-Task Data Construction}
\label{sec:data-construction}
Figure~\ref{fig:overview}(a) summarizes data construction.
We use the open-source CodeV-R1-dataset \citep{zhu2025codevr1}
for CoT supervision and its specification--RTL pairs as the source
for multi-task TIR data.
We use DeepSeek-V3.2 as the teacher for quality scoring and TIR
trajectory distillation.
The teacher scores pairs along four dimensions:
specification--implementation alignment, hardware-semantic precision,
behavioral reasoning value, and the ability to derive discriminating checks.
Pairs above the quality threshold form a shared source pool
(scoring prompt and threshold in Appendix~\ref{app:data-details}).
Each distilled trajectory retains the teacher's reasoning, tool requests,
actual observations, subsequent revisions, and final RTL.
Below, we describe the data construction and trajectory acceptance
criteria for each task.
\paragraph{Generation: constructing an implementation from a specification.}
Given a specification, the teacher writes candidate RTL and a simulation
testbench, then uses simulation diagnostics to revise the design as needed.
We retain trajectories whose final RTL passes the teacher-authored testbench
and is synthesizable by Yosys.

\paragraph{Completion: recovering logic in an existing design.}
An LLM identifies core logic in the reference RTL, which we mask while
retaining the surrounding design and interfaces as context.
The teacher fills in the missing logic, checks the completed design
through the simulation interface, and revises it using the returned feedback.
We retain a trajectory only if its final RTL passes equivalence checking
against the unmasked reference RTL, using the checker described in
Section~\ref{sec:setup}.
Generation and completion thus share a simulation interface but impose
different implementation constraints.
\paragraph{Optimization: improving a design while preserving its behavior.}
We run EDA checks on the reference designs in the shared source pool
and select those with lint warnings or relatively high synthesis-based
implementation costs.
These designs define tasks with explicit lint or implementation-metric
objectives.
The teacher iteratively modifies each design using the corresponding
diagnostics and equivalence-checking feedback.
We retain a trajectory only if its final RTL passes equivalence checking
against the original design and meets the requested improvement:
making a non-clean design lint-clean, or exceeding the prescribed
improvement threshold for the target synthesis-based PPA proxy.
These task-specific acceptance criteria yield 13,480 TIR trajectories.
Appendix~\ref{app:data-details} provides further data details.
\paragraph{RL task selection.}
Let $\mathcal D_{\mathrm{TIR}}$ denote the task pool corresponding to
the retained TIR demonstrations.
We sample five attempts per task using the two-stage SFT model and let
$c_{\mathrm{SFT}}(x)\in\{0,\ldots,5\}$ count successful attempts.
Functional correctness is checked by comparing the student's final RTL
against the reference RTL with the same equivalence checker used for
completion and optimization data.
Optimization success additionally requires the specified improvement.
We exclude tasks solved in all five attempts:
\begin{equation}
\mathcal D_{\mathrm{RL}}
=\{x\in\mathcal D_{\mathrm{TIR}}:c_{\mathrm{SFT}}(x)<5\}.
\end{equation}
This retains approximately 8K tasks for fresh policy rollouts,
excluding tasks already solved consistently by the student.
\FloatBarrier
\Needspace{20\baselineskip}
\subsection{Staged Supervision for a Single Policy}
\label{sec:staged-supervision}

\begin{wraptable}{r}{0.5\textwidth}
\vspace{-\intextsep}
\setlength{\abovecaptionskip}{0pt}
\centering
\caption{Controlled CoT/TIR supervision on 8B. Tool use is the share of
trajectories containing a tool request; Avg. tool turn is the mean number of tool-call turns in trajectories. Token counts denote the total number of chat-template tokens processed across all training epochs; for sequential configurations, the budget is shared across both stages.}
\label{tab:curriculum}
\papertablestyle
\begin{papertablefit}
\begin{tabular}{@{}lrrrr@{}}
\toprule
\textbf{Configuration} & \textbf{Epochs} & \makecell{\textbf{CVDP}\\\textbf{P@1 (\%)}} &
\makecell{\textbf{Tool}\\\textbf{use}} & \makecell{\textbf{Avg.}\\\textbf{tool turn}} \\
\midrule
Base model & 0 & 17.2 & 1.6\% & 0.0 \\
CoT (450M tokens) & 3 & 23.8 & 0.0\% & 0.0 \\
TIR (450M tokens) & 3 & 24.8 & 96.0\% & 4.0 \\
CoT (900M tokens) & 3 & 26.5 & 0.0\% & 0.0 \\
TIR (900M tokens) & 3 & 28.5 & 98.0\% & 3.7 \\
Naive mix (900M tokens) & 3 & 28.2 & 79.8\% & 2.7 \\
TIR$\rightarrow$CoT (900M tokens) & 3 & 25.5 & 0.0\% & 0.0 \\
\rowcolor{PaperBlue!8}
CoT$\rightarrow$TIR (900M tokens) & 3 & \textbf{29.5} & 97.4\% & 4.3 \\
\bottomrule
\end{tabular}
\end{papertablefit}
\end{wraptable}

We use CoT-to-TIR staged supervision (Figure~\ref{fig:overview}(b)).
The CoT stage reinforces RTL syntax, hardware knowledge, and reasoning from
specifications to implementations through direct specification--reasoning--code
examples. Building on this initialization, TIR supervision teaches the same
model to construct EDA requests and use the returned feedback to check and
revise its implementation.

Table~\ref{tab:curriculum} compares supervision configurations at matched template-token budgets.
The total budgets are 450M or 900M chat-template tokens across all training epochs. For the 900M-token mixed and sequential configurations, CoT and TIR each contribute 450M tokens; sequential configurations differ only in stage order.
CoT$\to$TIR yields 97.4\% tool use, versus 0.0\% for TIR$\to$CoT, supporting this order for retaining tool-use behavior.
At the same 900M-token budget, CoT$\to$TIR reaches 29.5\% Pass@1, exceeding TIR-only training (28.5\%) and naive CoT/TIR mixing (28.2\%).
These results support CoT-to-TIR staged supervision for direct RTL generation and feedback-guided revision; Section~\ref{sec:feedback} evaluates both abilities separately using the full-data checkpoints.
\par\WFclear

\subsection{Pattern-Aware Policy Optimization}
\label{sec:papo}

We introduce Pattern-Aware Policy Optimization (PAPO) to distinguish trajectories
with the same final correctness but different EDA-checking histories. Here, a
checking pattern characterizes whether a trajectory succeeds on its first check,
succeeds after feedback-driven revision, or lacks a successful task-appropriate
check. Starting from the two-stage SFT model, PAPO combines correctness-gated,
pattern-conditioned rewards with task-grouped policy updates within GRPO
\citep{shao2024deepseekmath} (Figure~\ref{fig:overview}(c)).

\begingroup
\predisplaypenalty=0
At update $k$, let $\mathcal B_k^{\mathrm F}$ and $\mathcal B_k^{\mathrm O}$
contain the retained generation/completion and optimization trajectories,
respectively. For trajectory $\tau_i$ sampled for task $x_i$, PAPO uses the
following objective and reward:
\begin{equation}
\mathcal J_{\mathrm{PAPO}}(\theta)
=\underbrace{
\sum_{g\in\{\mathrm F,\mathrm O\}}
\frac{\omega_{g,k}}{|\mathcal B_k^g|}
\sum_{i\in\mathcal B_k^g}
}_{\text{task-grouped updates}}
\frac{1}{|\mathcal M_i|}\sum_{t\in\mathcal M_i}c_{i,t}
\min\!\left\{
\rho_{i,t}\widehat A_i,\;
\operatorname{clip}\!\left(\rho_{i,t},1-\varepsilon_-,1+\varepsilon_+\right)
\widehat A_i\right\},
\label{eq:papo}
\end{equation}
\begin{equation}
\widehat A_i=
\frac{R_{x_i}(\tau_i)-\bar R_{x_i}}{\sigma_{x_i}+\epsilon},
\qquad
R_x(\tau)=\underbrace{V_x(y)\left[1+\delta_{p_x(\tau)}\right]}
_{\substack{\text{correctness-gated,}\\\text{pattern-conditioned reward}}}.
\label{eq:reward}
\end{equation}
Equation~\eqref{eq:reward} connects the checking pattern to the advantage
used in Equation~\eqref{eq:papo}. Definitions of the standard policy-update
quantities are provided in Appendix~\ref{app:training-config}.

\endgroup

\paragraph{Pattern-aware trajectory rewards.}
\label{sec:pattern}
Binary final-correctness rewards cannot distinguish EDA-checking histories. We therefore condition rewards on the EDA-checking pattern $p_x(\tau)$ through adjustment $\delta_{p_x(\tau)}$, gated by final verification (Equation~\eqref{eq:reward}). A checking round comprises the checks needed to assess a candidate against its task objective. \emph{Immediate Self-check Resolution (ISR)} succeeds in one task-appropriate checking round; \emph{Feedback-driven Refinement Success (FRS)} uses multiple checking rounds and ends with a successful self-check; \emph{Unsuccessful Self-checking (US)} covers missing, incomplete, or unsuccessful final checks. Self-check success is distinct from independent final verification. Independent final verification requires functional correctness against the reference and any specified optimization improvement for $V_x(y)=1$; otherwise reward is zero. These rewards distinguish EDA-checking histories among final-correct solutions, favoring concise successful checking while retaining positive rewards for successful revision. Appendix~\ref{app:patterns} gives matching rules and reward values.

\paragraph{Task grouping.}
Random sampling from the imbalanced RL pool may bias training toward more frequent task types.
We therefore use task grouping to maintain prescribed proportions among the retained trajectories in each update, continuing sampling until each group meets its quota. Generation and completion share a functional-correctness objective and a simulation interface; optimization adds an improvement objective and uses synthesis-analysis and equivalence-checking interfaces. Sampling quotas keep the retained optimization share $\alpha_k=|\mathcal B_k^{\mathrm O}|/(|\mathcal B_k^{\mathrm F}|+|\mathcal B_k^{\mathrm O}|)$ near the target $\alpha$, up to integer rounding. In Equation~\eqref{eq:papo}, $\omega_{\mathrm F,k}=1-\alpha_k$ and $\omega_{\mathrm O,k}=\alpha_k$ express these sample proportions in the batch average; they are not applied again.

\section{Experiments}
\label{sec:experiments}
\begingroup
\setlength{\textfloatsep}{10pt plus 2pt minus 2pt}
\setlength{\dbltextfloatsep}{8pt plus 2pt minus 2pt}
\setlength{\intextsep}{8pt plus 2pt minus 2pt}
\ifdefined\compactdraft\else
\ifdefined\compactdraft
\begin{table}[H]
\else
\begin{table}[!t]
\fi
\centering
\caption{CVDP non-agentic code-generation results (\%). Entries are
Pass@1/Pass@5; Overall is weighted by category size. Size denotes total model parameters; -- indicates undisclosed size. Tool marks whether EDA
tools are available. Best and second-best results for each metric are shown in
\textbf{bold} and \underline{underlined}, respectively; ties share the same
formatting. $\dagger$ marks the tool-using distillation teacher; ChipMATE uses its V+P cross-verification workflow.}
\label{tab:cvdp-main}
\begingroup
\footnotesize
\setlength{\tabcolsep}{3.5pt}
\renewcommand{\arraystretch}{1.04}
\newcommand{\vendoricon}[1]{\raisebox{-0.20ex}{\includegraphics[height=1.38ex]{figures/vendor_icons/#1.pdf}}\hspace{0.32em}}
\newcommand{\firstbest}[1]{\textbf{#1}}
\newcommand{\secondbest}[1]{\underline{#1}}
\newcommand{\modelseparator}{%
  \arrayrulecolor{black!18}\specialrule{0.25pt}{0.8pt}{0.8pt}\arrayrulecolor{black}}
\resizebox{\linewidth}{!}{%
\begin{tabular}{@{}lcccccccc@{}}
\toprule
\textbf{Model} & \textbf{Size} & \textbf{Tool} & \textbf{Overall} &
CID002 & CID003 & CID004 & CID007 & CID016 \\
\midrule
\multicolumn{9}{@{}l}{\emph{General-purpose frontier models}} \\
\multirow{2}{*}{\vendoricon{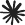}Claude-Sonnet-4.5}
 & \multirow{2}{*}{--} & \xmark & 46.6/59.6 & \secondbest{44.3}/53.2 & 48.2/61.5 & 49.1/60.0 & 39.5/62.5 & 53.7/68.6 \\
 &  & \cmark & \firstbest{51.5}/63.6 & \firstbest{46.8}/55.3 & \firstbest{56.9}/\firstbest{66.7} & \firstbest{54.2}/61.8 & 43.5/70.0 & 57.1/74.3 \\
\modelseparator
\multirow{2}{*}{\vendoricon{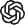}GPT-5.2}
 & \multirow{2}{*}{--} & \xmark & 41.5/57.6 & 36.6/50.0 & 45.1/57.7 & 42.9/61.8 & 35.0/57.5 & 52.0/71.4 \\
 &  & \cmark & \firstbest{51.5}/63.9 & 42.8/55.3 & \secondbest{51.8}/61.5 & \secondbest{50.9}/61.8 & \firstbest{63.5}/\firstbest{82.5} & 61.7/74.3 \\
\modelseparator
\multirow{2}{*}{\vendoricon{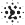}GLM-4.7}
 & \multirow{2}{*}{355B} & \xmark & 40.1/61.3 & 37.7/56.4 & 38.2/62.8 & 34.9/50.9 & 37.0/65.0 & \secondbest{62.3}/\secondbest{82.9} \\
 &  & \cmark & 41.5/62.3 & 38.7/\secondbest{57.4} & 44.9/\firstbest{66.7} & 38.5/56.4 & 36.0/57.5 & 52.6/80.0 \\
\modelseparator
\multirow{2}{*}{\vendoricon{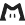}LongCat-Flash-Thinking}
 & \multirow{2}{*}{560B} & \xmark & 38.1/57.0 & 34.7/51.1 & 38.5/60.3 & 33.5/49.1 & 33.5/60.0 & 58.9/74.3 \\
 &  & \cmark & 39.5/58.6 & 33.0/50.0 & 43.3/60.3 & 34.9/54.5 & 38.0/67.5 & 57.1/74.3 \\
\modelseparator
\multirow{2}{*}{\vendoricon{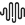}MiniMax-M2.1}
 & \multirow{2}{*}{230B} & \xmark & 33.3/53.6 & 29.8/50.0 & 39.0/60.3 & 34.5/56.4 & 21.0/40.0 & 42.3/60.0 \\
 &  & \cmark & 37.9/56.3 & 34.9/53.2 & 42.8/57.7 & 36.4/56.4 & 32.5/57.5 & 43.4/60.0 \\
\modelseparator
\multirow{2}{*}{\vendoricon{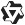}Qwen3-Max-Preview}
 & \multirow{2}{*}{--} & \xmark & 34.6/46.4 & 29.8/40.4 & 41.0/53.8 & 34.2/43.6 & 24.5/40.0 & 45.7/57.1 \\
 &  & \cmark & 33.7/49.0 & 29.4/40.4 & 39.2/55.1 & 30.5/40.0 & 28.5/52.5 & 44.0/68.6 \\
\modelseparator
\multirow{2}{*}{\vendoricon{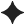}Gemini-2.5-Pro}
 & \multirow{2}{*}{--} & \xmark & 39.1/57.9 & 35.7/52.1 & 36.9/55.1 & 35.6/56.4 & 32.0/57.5 & \firstbest{66.3}/\secondbest{82.9} \\
 &  & \cmark & 38.2/57.9 & 33.2/46.8 & 37.9/60.3 & 40.0/61.8 & 27.5/52.5 & 61.7/\secondbest{82.9} \\
\midrule
\multicolumn{9}{@{}l}{\emph{RTL-specialized models}} \\
\multirow{2}{*}{CodeV-R1 {\scriptsize\citep{zhu2025codevr1}}}
 & \multirow{2}{*}{7B} & \xmark & 31.7/46.4 & 32.3/47.9 & 42.3/59.0 & 29.1/45.5 & 17.0/25.0 & 26.9/40.0 \\
 &  & \cmark & 23.2/42.7 & 20.9/40.4 & 34.6/53.8 & 27.3/50.9 & 8.0/22.5 & 15.4/34.3 \\
\modelseparator
\multirow{2}{*}{ChipSeek-R1 {\scriptsize\citep{chen2026chipseek}}}
 & \multirow{2}{*}{7B} & \xmark & 20.4/33.1 & 20.4/31.9 & 27.7/39.7 & 20.0/36.4 & 14.0/25.0 & 12.0/25.7 \\
 &  & \cmark & 21.5/32.8 & 21.5/29.8 & 27.7/42.3 & 24.4/40.0 & 15.0/22.5 & 10.9/20.0 \\
\modelseparator
\multirow{2}{*}{SiliconMind-V1 {\scriptsize\citep{chen2026siliconmind}}}
 & \multirow{2}{*}{8B} & \xmark & 33.7/51.3 & 30.6/46.8 & 41.5/59.0 & 33.5/54.5 & 22.0/35.0 & 38.3/60.0 \\
 &  & \cmark & 33.7/52.0 & 31.3/46.8 & 40.5/59.0 & 34.9/52.7 & 22.0/40.0 & 36.6/62.9 \\
\modelseparator
\multirow{2}{*}{InCoder {\scriptsize\citep{yang2026incoder}}}
 & \multirow{2}{*}{32B} & \xmark & 24.4/46.0 & 25.1/45.7 & 24.4/48.7 & 27.3/41.8 & 14.0/32.5 & 29.7/62.9 \\
 &  & \cmark & 28.1/47.7 & 30.0/47.9 & 31.0/53.8 & 28.7/47.3 & 16.0/30.0 & 29.1/54.3 \\
\modelseparator
ChipMATE {\scriptsize\citep{yu2026chipmate}}
 & \makecell{V: 9B\\P: 9B} & \cmark & 23.8/36.1 & 23.8/33.0 & 30.3/43.6 & 24.4/38.2 & 14.5/25.0 & 19.4/37.1 \\
\midrule
\multicolumn{9}{@{}l}{\emph{Distillation teacher}} \\
\multirow{2}{*}{\vendoricon{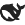}DeepSeek-V3.2$^\dagger$}
 & \multirow{2}{*}{671B} & \xmark & 36.8/57.3 & 34.5/54.3 & 38.7/60.3 & 36.4/49.1 & 32.0/57.5 & 45.1/71.4 \\
 &  & \cmark & 46.7/\secondbest{65.2} & 38.5/55.3 & 48.2/64.1 & 44.4/\secondbest{65.5} & \secondbest{60.0}/\secondbest{80.0} & 53.7/77.1 \\
\midrule
\multicolumn{9}{@{}l}{\emph{Our models}} \\
\vendoricon{qwen}Qwen3 (base) & 8B & \xmark & 18.9/36.1 & 14.3/30.9 & 17.4/34.6 & 22.9/38.2 & 18.5/37.5 & 29.1/48.6 \\
\vendoricon{qwen}RTL-Interact (SFT) & 8B & \cmark & 39.5/60.6 & 33.6/48.9 & 46.2/64.1 & 37.1/\secondbest{65.5} & 44.5/67.5 & 38.3/68.6 \\
\rowcolor{PaperBlue!8}
\vendoricon{qwen}\textbf{RTL-Interact (RL)} & 8B & \cmark & 44.9/63.2 & 37.0/56.4 & 47.7/61.5 & 46.5/\secondbest{65.5} & 52.5/77.5 & 48.6/65.7 \\
\modelseparator
\vendoricon{qwen}Qwen3 (base) & 32B & \xmark & 27.2/47.7 & 22.8/41.5 & 31.5/47.4 & 28.0/49.1 & 26.5/52.5 & 29.1/57.1 \\
\vendoricon{qwen}RTL-Interact (SFT) & 32B & \cmark & 42.5/64.6 & 38.9/\secondbest{57.4} & 45.1/61.5 & 39.6/63.6 & 44.5/\firstbest{82.5} & 48.0/71.4 \\
\rowcolor{PaperTeal!10}
\vendoricon{qwen}\textbf{RTL-Interact (RL)} & 32B & \cmark & \secondbest{48.7}/\firstbest{68.9} & 40.6/\firstbest{60.6} & 50.5/\secondbest{65.4} & 48.0/\firstbest{69.1} & 56.0/\secondbest{80.0} & 58.9/\firstbest{85.7} \\
\bottomrule
\end{tabular}
}%
\endgroup
\ifdefined\compactdraft
\end{table}
\else
\end{table}
\fi

\fi
\noindent\begin{minipage}[b]{\linewidth}
This section aims to answer the following research questions:
\par
\setlength{\parskip}{0pt}
\noindent \textbf{RQ1 (Overall Performance):} How effective is the RTL-Interact training framework across RTL benchmarks?\par
\noindent \textbf{RQ2 (PAPO Ablation):} Do pattern-aware rewards and task grouping improve RTL performance?\par
\noindent \textbf{RQ3 (Training Stages):} How do the training stages affect RTL generation and feedback-guided revision?\par
\noindent \textbf{RQ4 (Interaction Efficiency):} How does RL change tool use and response length?\par
\end{minipage}\par

\subsection{Evaluation Setup}

\paragraph{Benchmarks.}
We use CVDP's non-agentic split \citep{pinckney2025cvdp}: 302 problems covering completion, specification-to-RTL, modification, optimization, and debugging. VerilogEval v2 \citep{liu2023verilogeval,pinckney2024revisiting} provides 156 problems each for code completion (CC) and specification-to-RTL (SR); RTLLM v2 \citep{lu2024rtllm,liu2024openllm} contains 50 specification-to-RTL tasks.

\paragraph{Baselines.}
We select CodeV-R1-7B, ChipSeek-R1-Qwen2.5, SiliconMind-V1-8B, and InCoder-32B as RTL-specialized baselines and include ChipMATE's V+P workflow on CVDP. We include the DeepSeek-V3.2 teacher \citep{deepseek2025v32}, Qwen3-8B/32B bases \citep{yang2025qwen3}, and our SFT checkpoints. Table~\ref{tab:cvdp-main} also compares seven general-purpose frontier models selected around the teacher's release; Appendix~\ref{app:eval-baselines} lists them and their references.

\paragraph{Evaluation protocol.}
Except for ChipMATE's V+P workflow, baselines are evaluated with and without the same EDA interfaces. We use five responses per problem at temperature 0.7 and top-$p$ 0.95, reporting Pass@1/Pass@5 and category-weighted CVDP scores. All four RTL-specialized baselines cover both tool settings and VerilogEval v2 splits. ChipSeek uses the released Qwen2.5 checkpoint under our shared protocol. Further training and evaluation details are provided in Appendix~\ref{app:repro}.

\subsection{Main Results (RQ1)}
\label{sec:main-results}
\ifdefined\compactdraft

\fi

\ifdefined\compactdraft
\begin{table}[H]
\else
\begin{table}[!t]
\fi
\centering
\caption{VerilogEval v2 and RTLLM v2 results (\%). CC and SR denote code completion and specification-to-RTL.
Tools indicate EDA-tool availability. Size denotes total model parameters. Best and second-best results in each metric are
\textbf{bold} and \underline{underlined}; ties at displayed precision share formatting.
$\dagger$ marks the distillation teacher.}
\label{tab:verilogeval}
\begingroup
\papertablestyle
\renewcommand{\arraystretch}{1.04}
\newcommand{\vendoricon}[1]{\raisebox{-0.20ex}{\includegraphics[height=1.38ex]{figures/vendor_icons/#1.pdf}}\hspace{0.32em}}
\newcommand{\firstbest}[1]{\textbf{#1}}
\newcommand{\secondbest}[1]{\underline{#1}}
\newcommand{\modelseparator}{%
  \arrayrulecolor{black!18}\specialrule{0.25pt}{0.8pt}{0.8pt}\arrayrulecolor{black}}
\begin{papertablefit}
\begin{tabular}{@{}lcccc@{\hspace{8pt}}ccc@{\hspace{8pt}}cc@{}}
\toprule
& & \multicolumn{3}{c}{\textbf{VerilogEval v2 CC}} & \multicolumn{3}{c}{\textbf{VerilogEval v2 SR}} & \multicolumn{2}{c}{\textbf{RTLLM v2}} \\
\cmidrule(lr){3-5}\cmidrule(lr){6-8}\cmidrule(l){9-10}
\textbf{Model} & \textbf{Size} & \textbf{P@1} & \textbf{P@5} & \textbf{Tools} & \textbf{P@1} & \textbf{P@5} & \textbf{Tools} & \textbf{P@1/P@5} & \textbf{Tools} \\
\midrule
\multicolumn{10}{@{}l}{\emph{RTL-specialized models}} \\
\multirow{2}{*}{CodeV-R1 {\scriptsize\citep{zhu2025codevr1}}}
 & \multirow{2}{*}{7B} & 54.7 & 71.2 & \xmark & 53.3 & 67.3 & \xmark & 64.4/78.0 & \xmark \\
 &  & 51.0 & 73.7 & \cmark & 63.3 & 78.2 & \cmark & 74.4/82.0 & \cmark \\
\modelseparator
\multirow{2}{*}{ChipSeek-R1 {\scriptsize\citep{chen2026chipseek}}}
 & \multirow{2}{*}{7B} & 64.4 & 71.2 & \xmark & 64.5 & 71.2 & \xmark & 74.0/82.0 & \xmark \\
 &  & 64.7 & 70.5 & \cmark & 54.4 & 70.5 & \cmark & 71.2/76.0 & \cmark \\
\modelseparator
\multirow{2}{*}{SiliconMind-V1 {\scriptsize\citep{chen2026siliconmind}}}
 & \multirow{2}{*}{8B} & 69.9 & 80.1 & \xmark & 69.2 & 80.8 & \xmark & 70.0/84.0 & \xmark \\
 &  & 62.7 & 80.1 & \cmark & 64.7 & 82.7 & \cmark & 54.8/76.0 & \cmark \\
\modelseparator
\multirow{2}{*}{InCoder {\scriptsize\citep{yang2026incoder}}}
 & \multirow{2}{*}{32B} & 62.2 & 73.1 & \xmark & 61.5 & 74.4 & \xmark & 59.2/76.0 & \xmark \\
 &  & 62.1 & 78.2 & \cmark & 66.0 & 78.2 & \cmark & 66.0/84.0 & \cmark \\
\midrule
\multicolumn{10}{@{}l}{\emph{Distillation teacher}} \\
\multirow{2}{*}{\vendoricon{deepseek}DeepSeek-V3.2$^\dagger$}
 & \multirow{2}{*}{671B} & 76.5 & 85.9 & \xmark & 71.5 & 85.9 & \xmark & 75.6/\secondbest{86.0} & \xmark \\
 &  & 67.9 & 84.6 & \cmark & \firstbest{78.8} & \firstbest{87.8} & \cmark & 68.8/82.0 & \cmark \\
\midrule
\multicolumn{10}{@{}l}{\emph{Our models}} \\
\vendoricon{qwen}Qwen3 (base) & 8B & 46.5 & 62.8 & \xmark & 53.7 & 69.2 & \xmark & 45.2/62.0 & \xmark \\
\vendoricon{qwen}RTL-Interact (SFT) & 8B & 73.8 & \secondbest{86.5} & \cmark & 72.4 & 84.6 & \cmark & 76.4/\secondbest{86.0} & \cmark \\
\rowcolor{PaperBlue!8}
\vendoricon{qwen}\textbf{RTL-Interact (RL)} & 8B & 75.4 & 83.3 & \cmark & 75.1 & 84.6 & \cmark & \firstbest{77.6}/\secondbest{86.0} & \cmark \\
\modelseparator
\vendoricon{qwen}Qwen3 (base) & 32B & 64.0 & 77.6 & \xmark & 70.3 & 83.3 & \xmark & 64.4/80.0 & \xmark \\
\vendoricon{qwen}RTL-Interact (SFT) & 32B & \secondbest{76.7} & \secondbest{86.5} & \cmark & 72.7 & \secondbest{86.5} & \cmark & 76.0/\firstbest{88.0} & \cmark \\
\rowcolor{PaperTeal!10}
\vendoricon{qwen}\textbf{RTL-Interact (RL)} & 32B & \firstbest{80.3} & \firstbest{87.2} & \cmark & \secondbest{77.2} & \secondbest{86.5} & \cmark & \secondbest{77.2}/\firstbest{88.0} & \cmark \\
\bottomrule
\end{tabular}
\end{papertablefit}
\endgroup
\end{table}

\noindent\textbf{RTL-Interact-32B surpasses its tool-enabled distillation teacher on CVDP.}
RTL-Interact-32B achieves 48.7\% Pass@1 and 68.9\% Pass@5, exceeding the tool-enabled DeepSeek-V3.2 teacher by 2.0\% and 3.7\%, respectively (Table~\ref{tab:cvdp-main}).
It also achieves the highest overall CVDP Pass@5 among the compared models.
On VerilogEval v2 CC, the 32B model reaches 80.3\% Pass@1 and 87.2\% Pass@5, exceeding the teacher in both tool settings (Table~\ref{tab:verilogeval}).

\noindent\textbf{Both RTL-Interact model sizes outperform the evaluated RTL-specialized baselines of comparable size in Pass@1 across all three benchmarks.}
On CVDP, RTL-Interact-8B reaches 44.9\%/63.2\% Pass@1/Pass@5, exceeding the 7B--8B RTL-specialized baselines, while the 32B model exceeds InCoder-32B. Both also lead their corresponding baselines on both VerilogEval v2 splits and RTLLM v2 (Tables~\ref{tab:cvdp-main}--\ref{tab:verilogeval}). The 32B model reaches 77.2\%/86.5\% Pass@1/Pass@5 on VerilogEval v2 SR. On RTLLM v2, 8B and 32B reach 77.6\%/86.0\% and 77.2\%/88.0\% Pass@1/Pass@5, respectively, exceeding all evaluated baselines in Pass@1. Relative to SFT, RL yields CVDP Pass@1 gains of 5.4\% and 6.2\% for 8B and 32B, with Pass@1 gains on VerilogEval v2 and RTLLM v2.

\subsection{Ablation of PAPO (RQ2)}
\label{sec:rl-ablation}

\noindent\textbf{Both pattern-aware rewards and task grouping contribute to RTL performance.}
We examine the two components of PAPO on RTL-Interact-8B
(Table~\ref{tab:rl-components}). The \emph{w/o pattern-aware reward}
variant replaces the trajectory reward with binary final-verification
reward while retaining task grouping. The \emph{w/o task grouping}
variant retains pattern-aware rewards but samples tasks randomly from the RL dataset, without enforcing task-group proportions among the trajectories retained for each update.

\ifdefined\compactdraft\begin{table}[H]\else\begin{table}[!htbp]\fi
\centering
\caption{RL component ablations for RTL-Interact-8B on CVDP. All configurations
use tools; entries are Pass@1/Pass@5 (\%). Bold marks the best value for each
metric, including ties.}
\label{tab:rl-components}
\papertablestyle
\begin{papertablefit}
\begin{tabular}{@{}lcccccc@{}}
\toprule
\textbf{Configuration} & \textbf{Overall} & \textbf{CID002} &
\textbf{CID003} & \textbf{CID004} & \textbf{CID007} & \textbf{CID016} \\
\midrule
\rowcolor{PaperBlue!8}
Full method & \textbf{44.9}/\textbf{63.2} & \textbf{37.0}/\textbf{56.4} &
47.7/61.5 & \textbf{46.5}/\textbf{65.5} &
\textbf{52.5}/\textbf{77.5} & \textbf{48.6}/\textbf{65.7} \\
w/o pattern-aware reward & 39.7/59.9 & 36.6/55.3 &
46.2/64.1 & 40.7/61.8 & 34.0/57.5 & 38.9/62.9 \\
w/o task grouping & 41.8/60.3 & 36.0/52.1 &
\textbf{47.9}/\textbf{65.4} & 44.0/\textbf{65.5} & 41.5/60.0 & 40.6/62.9 \\
\bottomrule
\end{tabular}
\end{papertablefit}
\end{table}

The full method leads in overall Pass@1/Pass@5, with gains of 5.2\%/3.3\% over \emph{w/o pattern-aware reward} and 3.1\%/2.9\% over \emph{w/o task grouping}. We use an optimization-task share of $\alpha=0.3$ among samples retained for each policy update; Appendix~\ref{app:task-mixture} compares task-group proportions.
\par\WFclear

\Needspace{6\baselineskip}
\subsection{Direct Generation and Feedback-Guided Revision (RQ3)}
\label{sec:feedback}
\noindent\begin{wraptable}{r}{0.50\textwidth}
\vspace{-\intextsep}
\setlength{\abovecaptionskip}{0pt}
\centering
\caption{Direct generation and single-shot revision success (\%) on 100 tasks.
All checkpoints are 8B.}
\label{tab:feedback}
\papertablestyle
\begin{papertablefit}
\begin{tabular}{@{}lcc@{}}
\toprule
\textbf{Checkpoint} & \makecell{\textbf{Direct}\\\textbf{generation}} &
\makecell{\textbf{With EDA}\\\textbf{feedback}}\\
\midrule
Base & 17.0 & 14.0\\
CoT-SFT & 23.0 & 19.0\\
CoT$\to$TIR SFT & \textbf{27.0} & 26.0\\
\rowcolor{PaperTeal!10}
\textbf{PAPO} & 26.0 & \textbf{35.0}\\
\bottomrule
\end{tabular}
\end{papertablefit}
\end{wraptable}
\textbf{Staged supervision improves generation and feedback-guided revision, while RL further strengthens revision.}
Constructing candidate RTL and revising it using EDA feedback are two important capabilities for tool-assisted RTL design. We evaluate these capabilities separately on 100 selected CVDP tasks spanning all five categories. In direct generation, each model receives only the original task description and generates one RTL solution. In feedback-guided revision, all models receive the same fixed failing implementation and its EDA diagnostic for each task, then produce one revised implementation. Neither setting permits tool calls, allowing us to assess generation and revision separately from tool-use decisions. We compare the Qwen3-8B base, full CoT, two-stage SFT, and PAPO checkpoints using the CVDP evaluator. Detailed experimental settings and procedures are provided in Appendix~\ref{app:eval-config}.

Two-stage SFT raises generation/revision success from Base's 17.0\%/14.0\% to 27.0\%/26.0\%, also exceeding CoT's 23.0\%/19.0\% (Table~\ref{tab:feedback}). Together with the matched-budget study, this supports staged supervision for both abilities. RL further raises revision success to 35.0\% while retaining comparable generation performance.

\par\WFclear
\ifdefined\compactdraft\begin{figure}[H]\else\begin{figure*}[t]\fi
\centering
\setlength{\parskip}{0pt}
\setlength{\abovecaptionskip}{4pt}
\begin{minipage}[t]{0.49\linewidth}
\centering
\begin{tikzpicture}
\begin{axis}[
 name=axis8,scale only axis,width=.63\linewidth,height=2.0cm,
 xlabel={RL step},ylabel={Pass@1 (\%)},
 xmin=1,xmax=729,enlarge x limits=false,ymin=35,ymax=51,xtick={1,200,400,600,729},
 ytick={35,40,45,50},axis y line*=left,axis x line*=bottom,
 tick label style={font=\scriptsize},label style={font=\small},
 yticklabel style={text=PaperBlue},ylabel style={text=PaperBlue},
 grid=major,grid style={black!9},clip=true,
]
\addplot[PaperBlue!35,line width=.55pt]
 table[x=step,y=value]{data/train_8b_pass.dat};
\addplot[PaperBlue,line width=1.15pt]
 table[x=step,y=value]{data/train_8b_pass_ma5.dat};
\end{axis}
\begin{axis}[
 at={(axis8.south west)},anchor=south west,
 scale only axis,width=.63\linewidth,height=2.0cm,
 xmin=1,xmax=729,enlarge x limits=false,ymin=11,ymax=22,ytick={12,15,18,21},
 axis y line*=right,axis x line=none,ylabel={Length (k tokens)},
 tick label style={font=\scriptsize},label style={font=\small},
 yticklabel style={text=PaperOrange},ylabel style={text=PaperOrange},
]
\addplot[PaperOrange!40,line width=.55pt] table[x=step,y expr=\thisrow{value}/1000]{data/train_8b_length.dat};
\addplot[PaperOrange,line width=1.15pt] table[x=step,y expr=\thisrow{value}/1000]{data/train_8b_length_ma5.dat};
\end{axis}
\node[font=\small\bfseries,anchor=south] at ([yshift=2mm]axis8.north) {RTL-Interact-8B};
\end{tikzpicture}
\end{minipage}\hfill
\begin{minipage}[t]{0.49\linewidth}
\centering
\begin{tikzpicture}
\begin{axis}[
 name=axis32,scale only axis,width=.63\linewidth,height=2.0cm,
 xlabel={RL step},ylabel={Pass@1 (\%)},
 xmin=1,xmax=499,enlarge x limits=false,ymin=35,ymax=51,xtick={1,100,200,300,400,499},
 ytick={35,40,45,50},axis y line*=left,axis x line*=bottom,
 tick label style={font=\scriptsize},label style={font=\small},
 yticklabel style={text=PaperBlue},ylabel style={text=PaperBlue},
 grid=major,grid style={black!9},clip=true,
]
\addplot[PaperBlue!35,line width=.55pt]
 table[x=step,y=value]{data/train_32b_pass.dat};
\addplot[PaperBlue,line width=1.15pt]
 table[x=step,y=value]{data/train_32b_pass_ma5.dat};
\end{axis}
\begin{axis}[
 at={(axis32.south west)},anchor=south west,
 scale only axis,width=.63\linewidth,height=2.0cm,
 xmin=1,xmax=499,enlarge x limits=false,ymin=11,ymax=22,ytick={12,15,18,21},
 axis y line*=right,axis x line=none,ylabel={Length (k tokens)},
 tick label style={font=\scriptsize},label style={font=\small},
 yticklabel style={text=PaperOrange},ylabel style={text=PaperOrange},
]
\addplot[PaperOrange!40,line width=.55pt] table[x=step,y expr=\thisrow{value}/1000]{data/train_32b_length.dat};
\addplot[PaperOrange,line width=1.15pt] table[x=step,y expr=\thisrow{value}/1000]{data/train_32b_length_ma5.dat};
\end{axis}
\node[font=\small\bfseries,anchor=south] at ([yshift=2mm]axis32.north) {RTL-Interact-32B};
\end{tikzpicture}
\end{minipage}
\par\vspace{0mm}
\begin{tikzpicture}
\draw[PaperBlue,line width=1.15pt] (0,0)--(6mm,0);
\node[right,font=\scriptsize] at (7mm,0) {Pass@1};
\draw[PaperOrange,line width=1.15pt] (29mm,0)--(35mm,0);
\node[right,font=\scriptsize] at (36mm,0) {Mean response length};
\end{tikzpicture}
\caption{RL training dynamics for 8B and 32B. Pale solid
lines show the series before smoothing.}
\label{fig:training-dynamics}
\ifdefined\compactdraft\end{figure}
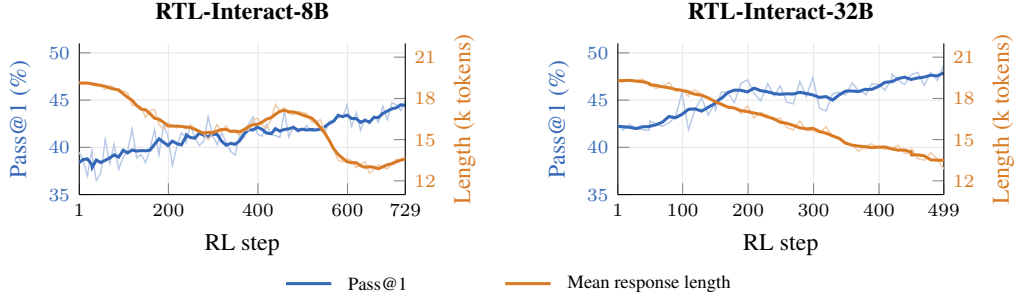\else\end{figure*}\fi

\Needspace{6\baselineskip}
\subsection{Interaction Efficiency (RQ4)}

\noindent\textbf{RL reduces tool calls and response length while improving RTL performance.}
Figure~\ref{fig:training-dynamics} shows that mean response length decreases by 28.8\% for 8B and 33.5\% for 32B over RL training.
On CVDP, RTL-Interact-8B's mean tool calls per trajectory decrease from 4.7 to 1.7, a 63.8\% reduction, while the truncation rate drops from 33.7\% to 15.2\%, a decrease of 18.5\%.
Detailed trajectory statistics are provided in Appendix~\ref{app:interaction-statistics}, Table~\ref{tab:trajectory}.
Together with the accuracy gains in Section~\ref{sec:main-results}, these results show that improved RTL performance does not require more tool calls or longer responses.
As a complementary observation, the first and total thinking lengths increase to $2.8\times$ and $1.5\times$ their SFT values, respectively, indicating that fewer tool calls coexist with more explicit thinking text.
\par

\section{Discussion and Conclusion}

We present \method, a post-training framework that combines multi-task data construction, CoT-to-TIR staged supervision, and PAPO to train a single model for RTL generation, autonomous EDA-tool use, and feedback-guided revision. Both model sizes outperform evaluated RTL-specialized baselines of comparable size in Pass@1 on CVDP, VerilogEval v2, and RTLLM v2; the 32B model also surpasses its tool-enabled teacher on CVDP. However, this study focuses on module-level RTL tasks and does not yet establish a protocol for repository-level development, where cross-module dependencies, build configurations, and system-level verification must be handled jointly. CoT and synthesized TIR data may retain errors, with TIR quality constrained by the teacher's capabilities and the coverage of automated checks. Future work will develop repository-level training environments and benchmarks and strengthen validation of synthesized trajectories.

\endgroup
\section*{Reproducibility Statement}

API and diagnostic archives retain responses and scores; diagnostic protocols
also preserve frozen inputs. Appendix~\ref{app:repro} documents data
construction, decontamination, training hyperparameters, evaluation settings,
and the information exposed through each tool interface.

\section*{AI Use Statement}
We used DeepSeek-V3.2 for quality scoring of specification-RTL pairs and for synthesizing tool-integrated reasoning (TIR) trajectories for generation, completion, and optimization tasks, as described in Section~\ref{sec:data-construction} and Appendix~\ref{app:data-details}.

AI tools also assisted literature search, writing, translation, LaTeX, and review, as well as parts of prompt design and code implementation.
The authors are responsible for experimental records, references, and claims.

\bibliography{references}
\bibliographystyle{iclr2027_conference}
\clearpage
\appendix
\section{Training and Evaluation Details}
\label{app:repro}

\subsection{Data Construction and Filtering}
\label{app:data-details}

We use 87,321 CoT examples from the open-source CodeV-R1-dataset \citep{zhu2025codevr1}.
We use its specification--RTL pairs as the source pool for multi-task TIR construction.
DeepSeek-V3.2 performs quality scoring and TIR trajectory distillation.

\paragraph{Quality-scoring prompt.}
Before constructing any of the three task types, the teacher acts as an
LLM judge for the shared instruction--RTL pool. Each pair receives a validity
label (\textsc{Pass}, \textsc{Review}, or \textsc{Reject}) and four integer
scores from 0 to 4:
\begin{enumerate}
  \item \emph{Specification--implementation alignment}: whether the
  specification defines the observable behavior and the RTL implements it;
  \item \emph{Hardware-semantic precision}: whether timing, state, finite
  width, signedness, and related hardware semantics must be handled precisely;
  \item \emph{Behavioral reasoning value}: whether multiple behavioral rules
  must be coordinated rather than mechanically transcribed; and
  \item \emph{Verifiability and error discrimination}: whether checks can be
  derived from the specification to distinguish correct behavior from
  plausible implementation errors.
\end{enumerate}
The prompt also requests evidence for every score, concrete specification or
code issues, up to two discriminating checks, difficulty, and confidence.
Only \textsc{Pass} pairs with a total score above 10 enter the shared candidate
pool for generation, completion, and optimization tasks; the score ranks
candidates and is not treated as a probability of correctness. The complete
prompt is included in the supplementary artifact.

\paragraph{TIR composition.}
Teacher outputs are executed and filtered according to each task objective.
For generation, the final design must pass simulation with the teacher's own
testbench and be synthesizable by Yosys. For completion, an LLM identifies
core logic in the reference RTL, which is masked while the surrounding design
is retained. The teacher's completed design must pass equivalence checking
against the unmasked reference. For optimization, EDA tools first screen the
source pool for designs with lint diagnostics or comparatively high synthesis
cost. A retained trajectory must preserve reference behavior through
equivalence checking and either make the design lint-clean or
improve the specified synthesis metric beyond the filtering threshold.
Both completion and optimization use the reference-based equivalence checker
reused from CodeV-R1.
The resulting 13,480 TIR trajectories comprise approximately 9K generation,
3K completion, and 1.5K optimization examples.

\paragraph{RL task pool.}
RL candidates come from TIR tasks whose teacher trajectories satisfy the
task-specific acceptance criteria above. We sample five attempts per task
from the two-stage SFT model and retain tasks for which fewer than five
attempts succeed. Functional correctness is determined by the same
equivalence checker, comparing final RTL with the reference design;
optimization also requires the specified improvement. The same checker
provides the final functional-correctness decision during RL.
This yields approximately 8K tasks:
about 7K generation/completion tasks and 1K optimization tasks.

\paragraph{Decontamination.}
Inspired by the $n$-gram decontamination approach of
\citet{guo2024deepseekcoder}, we use a 15-gram overlap scan between the training problems and the CVDP, VerilogEval v2, and RTLLM v2 test problems.
No overlapping training problems were identified by this procedure.

\subsection{Training Configuration}
\label{app:training-config}

\paragraph{Trajectory serialization and training targets.}
For training, the trajectory is serialized into model outputs interleaved
with tool observations. Tokens in the reasoning, tool requests, and final
RTL are model outputs; tool observations provide context for subsequent
generation but are not prediction targets. Section~\ref{sec:papo} assigns
a reward using the final RTL $y$ and the checking history recorded in the
tool requests and observations, then uses this reward to update the same
policy $\pi_\theta$.

The SFT stages use eight NVIDIA A100 GPUs, a batch size of 64, and a learning
rate of $10^{-5}$. CoT training runs for three epochs, followed by five epochs
of TIR training. The base checkpoints are Qwen3-8B and Qwen3-32B.

For multi-task RL, the reported 8B and 32B runs use a constant learning rate
of $10^{-6}$, 32 prompts per rollout batch, eight samples per prompt, and a
global batch size of 256. The retained generation/completion and optimization
prompt groups use per-update quotas targeting a 7:3 ratio, rounded to integer
group counts. Rollout temperature is 1.0 with a 30,000-token
response cap. GRPO uses clipping parameters 0.2/0.28 and no KL loss. Adam uses
$\beta_1=0.9$, $\beta_2=0.98$, $\epsilon=10^{-8}$, weight decay 0.01, and
seed 1234. Pattern-aware rewards are enabled for the reported RTL-Interact runs.
The binary-reward ablation uses the same initialization, data, optimizer,
rollout configuration, and number of training steps; only the
pattern-dependent reward weights are disabled.

\paragraph{Policy-update quantities.}
For trajectory $\tau_i$ sampled for task $x_i$, $\mathcal M_i$ contains the
indices of model-generated tokens, excluding tool observations. Here $t$ indexes tokens in the serialized trajectory, rather than the tool interactions indexed by $\ell$ in Section~\ref{sec:setup}. Let $u_{i,t}$
be the token at position $t$ and $h_{i,t}$ its preceding context, including the
task, previous model outputs, and tool observations. The policy ratio in
Equation~\eqref{eq:papo} is
\begin{equation}
\rho_{i,t}=\frac{\pi_\theta(u_{i,t}\mid h_{i,t})}
{\pi_{\mathrm{old}}(u_{i,t}\mid h_{i,t})},
\end{equation}
where $\pi_{\mathrm{old}}$ is the policy before the update. The parameters
$\varepsilon_-$ and $\varepsilon_+$ specify the lower and upper clipping
thresholds. We retain the fixed rollout correction
\begin{equation}
c_{i,t}=\operatorname{clip}\!\left(
\frac{\pi_{\mathrm{old}}(u_{i,t}\mid h_{i,t})}
{\pi_{\mathrm{roll}}(u_{i,t}\mid h_{i,t})},0,2\right),
\end{equation}
where $\pi_{\mathrm{roll}}$ generates the rollout and no gradient is propagated
through $c_{i,t}$.

Let $\mathcal I_x$ index the $G$ trajectories sampled for the same task $x$.
The reward statistics used to compute their advantages are
\begin{equation}
\bar R_x=\frac{1}{G}\sum_{j\in\mathcal I_x}R_x(\tau_j),
\qquad
\sigma_x=\operatorname{Std}_{j\in\mathcal I_x}\!\left[R_x(\tau_j)\right].
\end{equation}
The constant $\epsilon$ stabilizes division by the standard deviation.
Reward normalization is performed within the same task, rather than across
all tasks belonging to the same task group.

Token losses are averaged over model-generated tokens within each trajectory and then over retained trajectories. The same rollout correction and loss reduction are used in the binary-reward baseline.

\subsection{Evaluation Configuration}
\label{app:eval-config}

For the main benchmark evaluations, we cap model-generated responses at 32K tokens, excluding tool outputs.
We allow at most 6 tool calls per trajectory for functional tasks---generation, completion, modification, and bug fixing---and 10 for optimization tasks.
A response is counted as truncated if it reaches the response-length limit.

Unless stated otherwise, Pass@1/Pass@5 results use five samples per problem,
temperature 0.7, and top-$p$ 0.95. CVDP averages are weighted by category
sizes $(94,78,55,40,35)$, totaling 302. With $n$ samples and $c$ passing
samples, we use
\begin{equation}
\widehat{\mathrm{Pass@}k}=1-\frac{\binom{n-c}{k}}{\binom{n}{k}},\qquad n\geq k.
\end{equation}

\paragraph{Generation and revision protocol (Section~\ref{sec:feedback}).}
We freeze 100 tasks before comparative inference using source-balanced selection with seed 20260909, one task per name-based family, and preference for behavioral/lint failures rather than uniform random sampling.
Archived tool-use trajectories from GLM-4.6, MiniMax-M2, DeepSeek-V4-Flash, and a binary-reward 8B RL checkpoint (step 499) supply 25 failing candidates each. The completion/generation/modification/optimization/debugging counts are 24/28/11/20/17.
Candidates are intermediate RTL submissions that fail independent CVDP evaluation; diagnostics are reproduced with Icarus Verilog (47 compilation and 33 behavioral failures) or Verilator (20 lint failures).
Behavioral checks must also pass available archived CVDP-passing implementations.
Every revision model receives the same candidate, diagnostic, and archived model-authored testbench when present; hidden scoring tests remain unavailable. This diagnostic study uses temperature 1, top-$p$ 1, shared task-derived seeds, and initially one response per task/model/setting with a 32K-token generation budget.

\paragraph{General-purpose baselines.}
\label{app:eval-baselines}
The CVDP frontier baselines are Claude-Sonnet-4.5 \citep{anthropic2025sonnet45}, GPT-5.2 \citep{openai2025gpt52}, GLM-4.7 \citep{zai2025glm47}, LongCat-Flash-Thinking \citep{longcat2025thinking}, MiniMax-M2.1 \citep{minimax2025m21}, Qwen3-Max-Preview \citep{qwen2025maxpreview}, and Gemini-2.5-Pro \citep{gemini2025v25}. These checkpoints were selected around the teacher's release, reflecting the earlier start of our experiments.

\paragraph{Information boundary.}
\ifdefined\compactdraft\begin{table}[H]\else\begin{table}[ht]\fi
\centering
\caption{Five evaluated RTL task categories and four model-facing EDA
interfaces. Checkmarks denote tool access in our environment, rather than
restrictions on where an EDA tool can be useful. Equivalence checking uses
the original RTL supplied in an optimization task; hidden benchmark tests
are never exposed through these interfaces.}
\label{tab:tool-interfaces}
\papertablestyle
\begin{papertablefit}
\begin{tabular}{@{}lcccc@{}}
\toprule
\textbf{Task} & \textbf{Simulation} & \makecell{\textbf{Lint}\\\textbf{checking}} &
\makecell{\textbf{Synthesis}\\\textbf{analysis}} & \makecell{\textbf{Equivalence}\\\textbf{checking}}\\
\midrule
Generation & \cmark & --- & --- & ---\\
Completion & \cmark & --- & --- & ---\\
Modification & \cmark & --- & --- & ---\\
Debugging & \cmark & --- & --- & ---\\
Optimization & --- & \cmark & \cmark & \cmark\\
\bottomrule
\end{tabular}
\end{papertablefit}
\end{table}

Functional interaction requests contain model-generated RTL and a
model-generated testbench. The hidden/reference scoring testbench is not
provided by that interface. For optimization, the original design is part of
the task and is available for equivalence checking. Workflow systems may use
different checking information or budgets, so their published scores are not
inserted into our same-interface table.

\paragraph{Controlled supervision comparison.}
In Table~\ref{tab:curriculum}, the 450M-token budget corresponds to 150M chat-template
tokens per epoch for three epochs; the 900M-token budget doubles this total. The 17.2\% Base
result belongs to this controlled protocol and is not substituted for the
18.9\% tool-free main-table baseline.
Avg. tool turn reports the mean number of tool-call turns over all evaluated trajectories, including zero for trajectories without tool calls. Tool-call turns are distinct from the task-specific checking rounds defined below.

\Needspace{14\baselineskip}
\subsection{Interaction Statistics}
\label{app:interaction-statistics}
\begin{table}[H]
\centering
\caption{CVDP interaction statistics for RTL-Interact-8B before and after RL.
Tool calls and thinking lengths are means per trajectory; thinking lengths are measured in characters.
First and total thinking lengths refer to the first thinking segment and the sum of all thinking segments, respectively.
Truncation means reaching the response-length limit. Changes are reported in percentage points (pp) for truncation rate and as a relative percentage for tool calls.}
\label{tab:trajectory}
\papertablestyle
\begin{papertablefit}
\begin{tabular}{@{}lrrr@{}}
\toprule
\textbf{Metric} & \textbf{SFT} & \textbf{RL} & \textbf{Change} \\
\midrule
Tool calls & 4.7 & 1.7 & $-63.8\%$ \\
Truncation rate & 33.7\% & 15.2\% & $-18.5$ pp \\
First thinking length & 11,453 & 31,619 & $2.8\times$ \\
Total thinking length & 30,873 & 44,838 & $1.5\times$ \\
\bottomrule
\end{tabular}
\end{papertablefit}
\end{table}

\section{Checking Rounds and Pattern Rewards}
\label{app:patterns}

\paragraph{Checking rounds.}
The unit of checking reflects the task objective. For functional tasks,
one round consists of simulating the candidate with a model-authored
testbench. For optimization, one round combines a task-appropriate
optimization check (lint or synthesis analysis) with a self-equivalence
check against the original design. Thus two tool calls can constitute
one optimization checking round. In both cases, final correctness is
determined independently by the reference-based verifier.

\paragraph{Matching rules and weights.}
Let $n$ be the number of checking rounds and $p_j\in\{0,1\}$ indicate
whether round $j$ succeeds. For functional tasks, success means that the
model-authored simulation check passes. For optimization, it requires
successful task-appropriate optimization and equivalence checks.
Among task-appropriate checking histories, ISR requires $n=1$ and $p_1=1$;
FRS requires $n>1$ and $p_n=1$. An earlier successful check does not qualify
a trajectory for FRS if its final self-check fails. US covers histories
that do not meet either condition, including missing or incomplete checks
and the use of an optimization tool unrelated to the task objective.

For both functional and optimization tasks, let $p_x(\tau)\in\{\mathrm{ISR},\mathrm{FRS},\mathrm{US}\}$ denote the matched checking pattern. The pattern adjustment in Equation~\eqref{eq:reward} is
\begin{equation}
\delta_{p_x(\tau)}=
\begin{cases}
  +0.5, & \text{ISR},\\
  +0.2, & \text{FRS},\\
  -0.2, & \text{US}.
\end{cases}
\end{equation}
Thus $R_x(\tau)=V_x(y)[1+\delta_{p_x(\tau)}]$ gives final-correct
trajectories rewards of 1.5, 1.2, and 0.8, respectively. The value 1.0
is the base correctness reward before the pattern adjustment, not an
additional unpenalized US category.
For all tasks, a failed final verification gives zero reward regardless
of checking history.

\paragraph{Reward ablation.}
The binary ablation removes the pattern-conditioned successful rewards
while retaining the final correctness gate. It tests the whole reward
design, not each rule separately. There is no reported ablation against
a tuned length penalty, so the paper does not claim superiority to every
cost-aware reward.

\section{Task-Group Proportion Analysis}
\label{app:task-mixture}

We vary $\alpha$, the optimization-task share among samples retained for each policy update, with generation and completion sharing the remaining $1-\alpha$.
We compare $\alpha\in\{0.2,0.3,0.4\}$ for RTL-Interact-8B on CVDP.
Figure~\ref{fig:task-mixture} shows that $\alpha=0.3$ yields the highest overall Pass@1 and Pass@5 among the three tested settings, reaching 44.9\% and 63.2\%, respectively.
We use this proportion in the full method.

\begin{figure}[H]
\centering
\begin{minipage}[t]{0.48\textwidth}
\vspace{0pt}
\centering
\begin{tikzpicture}
\begin{axis}[
 scale only axis,width=.76\linewidth,height=2.8cm,
 xmin=.18,xmax=.42,ymin=38,ymax=68,
 xtick={.2,.3,.4},xticklabels={0.2,0.3,0.4},
 ytick={40,50,60},
 xlabel={Optimization share $\alpha$},ylabel={Pass rate (\%)},
 tick label style={font=\scriptsize},label style={font=\small},
 axis x line*=bottom,axis y line*=left,
 axis line style={black!45},tick style={black!45},
 ymajorgrids,grid style={black!9},
 legend columns=2,
 legend style={font=\scriptsize,draw=none,fill=none,
   at={(.5,1.03)},anchor=south,column sep=5pt},
 legend cell align=left,
 nodes near coords,
 nodes near coords style={font=\scriptsize,anchor=south,yshift=2pt,
   /pgf/number format/.cd,fixed,precision=1,fixed zerofill},
]
\addplot[PaperBlue,line width=1.1pt,mark=*,mark size=2.4pt,
 mark options={solid,fill=PaperBlue,draw=white,line width=.45pt}]
 table[x=alpha,y=pass]{data/mixture_ablation.dat};
\addlegendentry{Pass@1}
\addplot[PaperTeal,line width=1.1pt,mark=diamond*,mark size=3pt,
 mark options={solid,fill=PaperTeal,draw=white,line width=.45pt}]
 table[x=alpha,y=pass5]{data/mixture_ablation.dat};
\addlegendentry{Pass@5}
\end{axis}
\end{tikzpicture}
\captionof{figure}{CVDP Pass@1/Pass@5 (\%) versus the optimization share $\alpha$
among samples retained for each RL update; the functional share is $1-\alpha$.}
\label{fig:task-mixture}
\end{minipage}\hfill%
\begin{minipage}[t]{0.48\textwidth}
\vspace{0pt}
\centering
\captionof{table}{Endpoint results for generation-only RL on RTL-Interact-8B. Values are
Pass@1/Pass@5.}
\label{tab:generation-transfer}
\papertablestyle
\begin{papertablefit}
\begin{tabular}{@{}lcc@{}}
\toprule
\textbf{Task category} & \textbf{SFT} & \makecell{\textbf{Generation-only}\\\textbf{RL}}\\
\midrule
CID002 & 33.4/55.3 & 38.5/53.2\\
CID003 & 40.5/53.8 & 49.0/62.8\\
CID004 & 35.3/58.2 & 47.3/63.6\\
CID007 & 23.0/52.5 & 26.5/50.0\\
CID016 & 38.3/62.9 & 50.3/71.4\\
\rowcolor{PaperBlue!8}
Overall & 34.8/56.0 & 42.6/59.3\\
\bottomrule
\end{tabular}
\end{papertablefit}

\end{minipage}
\end{figure}

\section{Additional Generation-Only Transfer Results}
\label{app:transfer}

The generation-only run starts from its own SFT checkpoint. Pass@1 rises
from 34.8\% to 42.6\% overall, with gains of 5.1\%, 8.5\%, 12.0\%, 3.5\%, and 12.0\%
on completion, specification-to-RTL, modification, optimization, and debugging, respectively. This transfer pattern is consistent with benefits
across functional tasks sharing an interaction interface. Optimization
also changes the objective and correctness conditions, however, so this
comparison does not isolate the causal effect of using a different tool.
Optimization Pass@5 decreases from 52.5\% to 50.0\%, and completion Pass@5
decreases from 55.3\% to 53.2\%. All these endpoints are retained in Table~\ref{tab:generation-transfer}; no intermediate category trajectory is inferred from them.

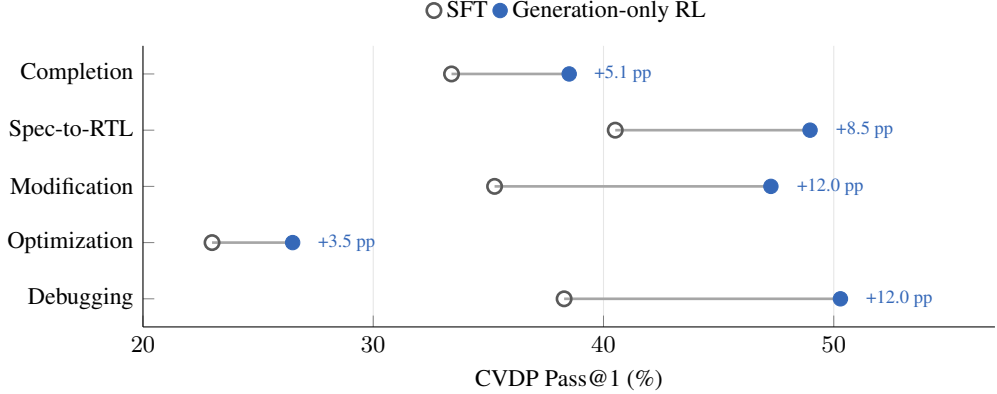
\begin{figure}[H]
\centering
\begin{tikzpicture}
\begin{axis}[
 width=.92\linewidth,height=5.3cm,
 xmin=20,xmax=57,ymin=.5,ymax=5.5,
 xlabel={CVDP Pass@1 (\%)},
 ytick={1,2,3,4,5},
 yticklabels={Debugging,Optimization,Modification,Spec-to-RTL,Completion},
 xtick={20,30,40,50},xmajorgrids,grid style={black!10},
 axis x line*=bottom,axis y line*=left,
 tick label style={font=\small},label style={font=\small},
 legend style={draw=none,font=\small,at={(.5,1.05)},anchor=south,legend columns=2},
]
\addplot[black!35,line width=1pt,forget plot] coordinates {(33.40,5)(38.51,5)};
\addplot[black!35,line width=1pt,forget plot] coordinates {(40.51,4)(48.97,4)};
\addplot[black!35,line width=1pt,forget plot] coordinates {(35.27,3)(47.27,3)};
\addplot[black!35,line width=1pt,forget plot] coordinates {(23.00,2)(26.50,2)};
\addplot[black!35,line width=1pt,forget plot] coordinates {(38.29,1)(50.29,1)};
\addplot[only marks,mark=o,mark size=2.7pt,black!65,line width=1pt]
 coordinates {(33.40,5)(40.51,4)(35.27,3)(23.00,2)(38.29,1)};
\addlegendentry{SFT}
\addplot[only marks,mark=*,mark size=2.7pt,PaperBlue]
 coordinates {(38.51,5)(48.97,4)(47.27,3)(26.50,2)(50.29,1)};
\addlegendentry{Generation-only RL}
\node[anchor=west,font=\scriptsize,text=PaperBlue] at (axis cs:39.2,5) {+5.1 pp};
\node[anchor=west,font=\scriptsize,text=PaperBlue] at (axis cs:49.7,4) {+8.5 pp};
\node[anchor=west,font=\scriptsize,text=PaperBlue] at (axis cs:48.0,3) {+12.0 pp};
\node[anchor=west,font=\scriptsize,text=PaperBlue] at (axis cs:27.2,2) {+3.5 pp};
\node[anchor=west,font=\scriptsize,text=PaperBlue] at (axis cs:51.0,1) {+12.0 pp};
\end{axis}
\end{tikzpicture}
\caption{Measured endpoint changes in the separate 8B generation-only run.
Segments connect SFT and RL results within a category; they are not
training-time curves. Labels show Pass@1 gains in percentage points (pp).
This run has a different SFT initialization from the full multi-task run.}
\label{fig:generation-transfer}
\end{figure}

\section{Evaluation Prompts}
\label{app:prompts}
We provide the common prompt templates used across all evaluated benchmarks, with and without tool access. \texttt{\{problem\_description\}} is replaced with the task description, and \texttt{\{task\_specific\_tool\_definitions\}} contains the tool interfaces available for the corresponding task. We display the message contents by role, omitting chat-template delimiters. Line wrapping is adjusted for readability. 

\subsection{Tool-Free Prompt}
\label{app:prompt-notool}
\begin{tcolorbox}[paperexample,title={Prompt 1: Tool-free evaluation}]
\textbf{System}
\begin{lstlisting}[style=paperliteral]
You are a Verilog RTL designer that only writes code using correct
Verilog syntax.
\end{lstlisting}
\textbf{User}
\begin{lstlisting}[style=paperliteral]
Question:
{problem_description}

Enclose your code with [BEGIN] and [DONE]. Only output the code snippet
and do NOT output anything else.

Answer:
\end{lstlisting}
\end{tcolorbox}
\subsection{Tool-Enabled Prompt}
\label{app:prompt-tool}
\begin{tcolorbox}[paperexample,title={Prompt 2: Tool-enabled evaluation},title after break={Prompt 2: Tool-enabled evaluation (continued)}]
\textbf{System}
\begin{lstlisting}[style=paperliteral]
You are a Verilog coding expert who will write Verilog code after careful consideration.
Your goal is to deliver functionally correct, synthesizable, and efficient RTL that meets the user's requirements.

Principles:
- Reason carefully about the specification, interfaces, and corner cases.
- Validate behavior and iterate as needed using any available tools.
- Iverilog Compatibility: Use robust Verilog-2001 or SystemVerilog-2012 constructs.
  - You should prefer `logic`, `always_ff`, and `always_comb`.
  - Avoid complex SystemVerilog features that have limited support in iverilog (e.g., specialized interfaces, program blocks, or nested struct assignments).
- Prefer clean, portable RTL: separate combinational/sequential logic, avoid unintended latches, avoid unsynthesizable constructs in design code.
- Use clear resets, parameterize widths when appropriate, and write readable, well-commented code.
- When a testbench is requested or required (e.g. by tools), follow these rules:
  1. Clearly state the expected behavior/results in the testbench (e.g. as comments).
  2. Ensure the testbench checks that simulation outputs exactly match the expected results; any mismatch means the design is still buggy.
  3. If the module has a reset signal, include proper reset testing in the testbench.
  4. Provide sufficient and diverse stimulus: cover normal cases, boundary conditions, and corner cases (e.g. 0, maximum values, negative numbers, sign-bit cases, etc.) instead of testing only a single scenario.
  5. Use $display for textual checks. If you emit errors, surface them clearly.
  6. DO NOT use `$monitor` or print signals every clock cycle. This will exceed the context limit.
After thinking, when you finally reach a conclusion, enclose the final verilog code in verilog code block within code tags. i.e.,
```verilog
module top_module(in, out, ...);
  ...
endmodule
```


# Tools

You may call one or more functions to assist with the user query.

You are provided with function signatures within <tools></tools> XML tags:
<tools>
{task_specific_tool_definitions}
</tools>

For each function call, return a json object with function name and arguments within <tool_call></tool_call> XML tags:
<tool_call>
{"name": <function-name>, "arguments": <args-json-object>}
</tool_call>
\end{lstlisting}
\textbf{User}
\begin{lstlisting}[style=paperliteral]
Design a functionally correct, synthesizable, and efficient Verilog module for the following problem.
Use any available tools during the process to validate and refine your design if you think it's needed.

The problem is as follows:
{problem_description}
\end{lstlisting}
\end{tcolorbox}

\section{Case Study: Response Trajectories}
\label{app:case-study}
Below are the original problem and response trajectories of the 8B Base, SFT, and RL models on the RTLLM v2 \texttt{freq\_divbyeven} task. Input prompts are omitted. Within each response turn, reasoning and tool-call contents are summarized and marked \texttt{[Summary]} inside their respective tags. Tool feedback and final answers are reproduced without summarization.

\begin{tcolorbox}[paperexample,title={Original problem: Even-frequency divider},title after break={Original problem: Even-frequency divider (continued)}]
\begin{lstlisting}[style=paperliteral,breakatwhitespace=false,literate={→}{{$\rightarrow$}}1 {-}{{-}}1]
Please act as a professional verilog designer.

Frequency divider that divides the input clock frequency by even numbers. This module generates a divided clock signal by toggling its output every specified number of input clock cycles.

Module name: 
    freq_diveven

Input ports:
    clk: Input clock signal that will be divided.
    rst_n: Active-low reset signal to initialize the module.

Output ports:
    clk_div: Divided clock output signal.

Implementation:
    The frequency divider uses a counter (`cnt`) to count the number of clock cycles. The `NUM_DIV` parameter specifies the division factor, which must be an even number.
    - When the reset signal (`rst_n`) is low, the counter and the divided clock signal (`clk_div`) are initialized to zero.
    - On each positive edge of the input clock (`clk`), if the counter is less than half of `NUM_DIV - 1`, the counter increments without changing the divided clock output.
    - When the counter reaches the specified limit, it resets to zero and toggles the `clk_div` output signal, effectively dividing the frequency of the input clock by the even number specified by `NUM_DIV`.

Counter:
    - The counter is a 4-bit register (`cnt`) that tracks the number of clock cycles.

Give me the complete code.
\end{lstlisting}
\end{tcolorbox}

\begin{tcolorbox}[paperexample,title={Base-8B},title after break={Base-8B (continued)}]
\begin{lstlisting}[style=paperliteral,breakatwhitespace=false,literate={→}{{$\rightarrow$}}1 {-}{{-}}1]
===== MODEL RESPONSE t0 =====

<think>
[Summary]
The model considers conflicting interpretations of the counter limit and repeatedly confuses the interval between output toggles with the full output period. It ultimately chooses to toggle at NUM_DIV - 1, incorrectly treating this as division by NUM_DIV.
</think>

[BEGIN]
module freq_diveven (
    input clk,
    input rst_n,
    output reg clk_div
);

parameter NUM_DIV = 4; // Must be even
reg [3:0] cnt;

always @(posedge clk or negedge rst_n) begin
    if (!rst_n) begin
        cnt <= 4'd0;
        clk_div <= 1'b0;
    end else begin
        if (cnt == NUM_DIV - 1) begin
            cnt <= 4'd0;
            clk_div <= ~clk_div;
        end else begin
            cnt <= cnt + 1;
        end
    end
end

endmodule
[DONE]
\end{lstlisting}
\end{tcolorbox}

\begin{tcolorbox}[paperexample,title={RTL-Interact-8B (SFT)},title after break={RTL-Interact-8B (SFT) (continued)}]
\begin{lstlisting}[style=paperliteral,breakatwhitespace=false,literate={→}{{$\rightarrow$}}1 {-}{{-}}1]
===== MODEL RESPONSE t0 =====

<think>
[Summary]
The model resolves the half-period interpretation: division by NUM_DIV requires toggling every NUM_DIV/2 input cycles. It selects CNT_MAX = NUM_DIV/2 - 1 with a 4-bit counter and active-low reset, then prepares an initial simulation.
</think>

<answer>
</answer>

<tool_call>
[Summary]
Function: verilog_simulator
Argument code: Submit the DUT using CNT_MAX = NUM_DIV/2 - 1 and a testbench for NUM_DIV = 4. Check output timing and reset initialization; print ALL_TESTS_PASSED when the error count is zero.
</tool_call>

===== TOOL RESPONSE t0_tool0.json =====

Test completed. Error count = 0
ALL_TESTS_PASSED
candidate.sv:70: $finish called at 220 (1s)

===== MODEL RESPONSE t1 =====

<think>
[Summary]
After the first self-check passes, the model plans broader parameter coverage and additional reset checks using multiple DUT instances.
</think>

<answer>
</answer>

<tool_call>
[Summary]
Function: verilog_simulator
Argument code: Submit the same divider logic with a testbench instantiating NUM_DIV = 2, 4, 6, and 8. Check the expected output periods and reset behavior; print ALL_TESTS_PASSED when no errors are detected.
</tool_call>

===== TOOL RESPONSE t1_tool0.json =====

Test completed. Error count = 0
ALL_TESTS_PASSED
candidate.sv:77: $finish called at 220 (1s)

===== MODEL RESPONSE t2 =====

<think>
[Summary]
After the multi-parameter self-check passes, the model revisits the specification and rechecks the threshold calculation, counter width, and reset behavior. It retains the same implementation and prepares another verification call before returning the final code.
</think>

<answer>
</answer>

<tool_call>
[Summary]
Function: verilog_simulator
Argument code: Submit the divider and another testbench for NUM_DIV = 4, repeating verification after the previous checks passed. Print ALL_TESTS_PASSED when the error count is zero.
</tool_call>

===== TOOL RESPONSE t2_tool0.json =====

Test completed. Error count = 0
ALL_TESTS_PASSED
candidate.sv:64: $finish called at 220 (1s)

===== MODEL RESPONSE t3 =====

<think>
[Summary]
The model prepares the final DUT-only implementation, including explanatory comments and omitting the testbench.
</think>

<answer>
```verilog
/**
 * Frequency divider that divides the input clock frequency by an even number.
 *
 * @param NUM_DIV  Division factor (must be an even number)
 * @param clk      Input clock signal
 * @param rst_n    Active-low reset (0 = initialize, 1 = normal operation)
 * @param clk_div  Divided clock output
 *
 * The module uses a 4-bit counter to count input clock cycles.  When the
 * counter reaches (NUM_DIV/2)-1, it resets and toggles the output, producing
 * a clock with period = NUM_DIV * Tclk.
 */
module freq_diveven #(
    parameter NUM_DIV = 4   // must be an even number
) (
    input  wire clk,
    input  wire rst_n,
    output reg  clk_div
);

    // Half of the division factor (integer division, NUM_DIV is even)
    localparam HALF_DIV = NUM_DIV / 2;
    // Counter maximum value
    localparam CNT_MAX  = HALF_DIV - 1;

    // 4-bit counter as specified
    reg [3:0] cnt;

    always @(posedge clk or negedge rst_n) begin
        if (!rst_n) begin
            // Active-low reset: clear counter and output
            cnt      <= 4'b0;
            clk_div  <= 1'b0;
        end else begin
            if (cnt == CNT_MAX) begin
                // Reached the limit: reset counter and toggle output
                cnt      <= 4'b0;
                clk_div  <= ~clk_div;
            end else begin
                // Normal counting, keep output unchanged
                cnt      <= cnt + 1;
            end
        end
    end

endmodule
```
</answer>
\end{lstlisting}
\end{tcolorbox}

\begin{tcolorbox}[paperexample,title={RTL-Interact-8B (RL)},title after break={RTL-Interact-8B (RL) (continued)}]
\begin{lstlisting}[style=paperliteral,breakatwhitespace=false,literate={→}{{$\rightarrow$}}1 {-}{{-}}1]
===== MODEL RESPONSE t0 =====

<think>
[Summary]
The model examines the half-period interpretation and integer division, then selects HALF = (NUM_DIV - 1)/2. It develops a testbench with multiple DUT instances to check division factors 2, 4, 6, and 8 in a single simulation call.
</think>

<answer>
</answer>

<tool_call>
[Summary]
Function: verilog_simulator
Argument code: Submit the DUT using HALF = (NUM_DIV - 1)/2 and a testbench instantiating NUM_DIV = 2, 4, 6, and 8. Measure output periods and compare them with the expected values in one simulation; print ALL_TESTS_PASSED when the error count is zero.
</tool_call>

===== TOOL RESPONSE t0_tool0.json =====

ALL_TESTS_PASSED
candidate.sv:88: $finish called at 220000 (1ps)

===== MODEL RESPONSE t1 =====

<think>
[Summary]
After receiving a successful self-check, the model prepares the final DUT-only implementation with the same threshold, omitting the testbench.
</think>

<answer>
```verilog
// Frequency divider that divides the input clock frequency by an even number.
module freq_diveven #(
    parameter NUM_DIV = 4          // Must be an even number
) (
    input  wire clk,               // Input clock
    input  wire rst_n,             // Active-low reset
    output wire clk_div            // Divided clock output
);
    reg [3:0] cnt;                 // 4-bit counter
    reg       clk_div_reg;         // Internal register for clk_div

    // HALF = (NUM_DIV - 1) / 2  (integer division)
    localparam HALF = (NUM_DIV - 1) / 2;

    always @(posedge clk or negedge rst_n) begin
        if (!rst_n) begin
            // Reset: clear counter and output
            cnt        <= 4'b0;
            clk_div_reg <= 1'b0;
        end else begin
            if (cnt < HALF) begin
                // Increment counter, keep output unchanged
                cnt <= cnt + 1'b1;
            end else begin
                // Counter reached the limit: reset it and toggle the output
                cnt        <= 4'b0;
                clk_div_reg <= ~clk_div_reg;
            end
        end
    end

    assign clk_div = clk_div_reg;
endmodule
```
</answer>
\end{lstlisting}
\end{tcolorbox}

\end{document}